\documentclass[fleqn]{2022AMSE}
\usepackage{graphicx}
\usepackage{subcaption}
\usepackage{booktabs}
\usepackage{multirow} 
\usepackage{enumitem} 

\usepackage{bm}

\begin{document}

\ensubject{Solid Dynamics}

\ArticleType{RESEARCH PAPER}
\Year{2025}
\Vol{41}
\DOI{10.1007/s10409-025--x}
\ArtNo{xxx}
\ReceiveDate{xxx}
\AcceptDate{xxx}
\OnlineDate{xxx}

\title{3D Trajectory Reconstruction of Moving Points Based on Asynchronous Cameras}{3D Trajectory Reconstruction of Moving Points Based on Asynchronous Cameras}

\author[1,2]{Huayu Huang}{}%
\author[1,2]{Banglei Guan}{guanbanglei12@nudt.edu.cn}
\author[1,2]{Yang Shang}{}
\author[1,2]{Qifeng Yu}{yuqifeng@nudt.edu.cn}

\AuthorMark{H. Huang}

\AuthorCitation{H. Huang, B. Guan, Y. Shang, and Q. Yu}

\address[1]{College of Aerospace Science and Engineering, National University of Defense Technology, Changsha 410073, Hunan, China}
\address[2]{Hunan Provincial Key Laboratory of Image Measurement and Vision Navigation, Changsha 410073, Hunan, China}

\contributions{Executive Editor: xxx}

\abstract{Photomechanics is a crucial branch of solid mechanics. The localization of point targets constitutes a fundamental problem in optical experimental mechanics, with extensive applications in various missions of UAVs. Localizing moving targets is crucial for analyzing their motion characteristics and dynamic properties. Reconstructing the trajectories of points from asynchronous cameras is a significant challenge. It encompasses two coupled sub-problems: trajectory reconstruction and camera synchronization. Present methods typically address only one of these sub-problems individually. This paper proposes a 3D trajectory reconstruction method for point targets based on asynchronous cameras, simultaneously solving both sub-problems. Firstly, we extend the trajectory intersection method to asynchronous cameras to resolve the limitation of traditional triangulation that requires camera synchronization. Secondly, we develop models for camera temporal information and target motion, based on imaging mechanisms and target dynamics characteristics. The parameters are optimized simultaneously to achieve trajectory reconstruction without accurate time parameters. Thirdly, we optimize the camera rotations alongside the camera time information and target motion parameters, using tighter and more continuous constraints on moving points. The reconstruction accuracy is significantly improved, especially when the camera rotations are inaccurate. Finally, the simulated and real-world experimental results demonstrate the feasibility and accuracy of the proposed method. The real-world results indicate that the proposed algorithm achieved a localization error of 112.95 m at an observation distance range of 15 $\sim$ 20 km.}


\setlength{\textheight}{23.6cm}
\thispagestyle{empty}

\maketitle
\setlength{\parindent}{1em}

\vspace{-1mm}
\begin{multicols}{2}

\section{Introduction}\label{sec1}

Photomechanics is a crucial branch of solid mechanics \cite{JW1}. It has significant applications in various fields such as aviation, aerospace, and military \cite{JW2,JW3,JW4,JW5}. In the fields of photomechanics, the use of triangulation for target measurement holds significant research and application value. For instance, in various unmanned aerial \Authorfootnote vehicles (UAVs) applications, such as damage detection, smart cities, and target reconnaissance, the localization of targets is of crucial importance \cite{JW6,JW7,JW8,JW9}. Almost all of these applications require rapid, high-precision localization of the target of interest. Localizing moving targets is especially crucial for analyzing their motion and dynamic characteristics. The localization of targets by UAVs mainly relies on their onboard electro-optical platforms, particularly the cameras.
\begin{figure*}[t] 
\centering
\subfloat[Triangulation of a static point]{\includegraphics[width=2.5in]{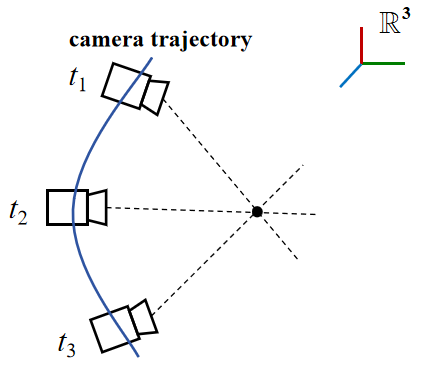}%
\label{fig1a}}
\hfil
\subfloat[Triangulation of a moving point]{\includegraphics[width=2.5in]{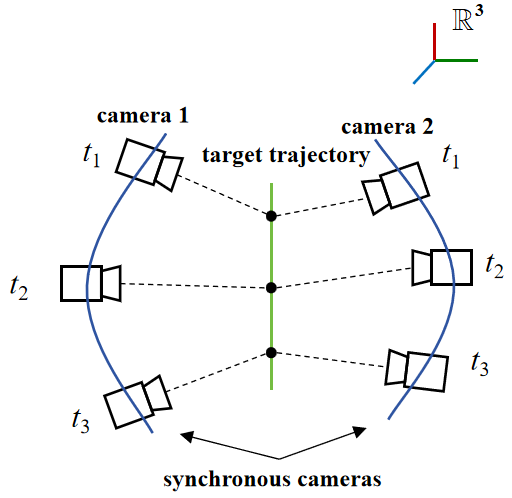}%
\label{fig1b}}
\caption{Illustration of triangulation.}
\label{fig1}
\end{figure*}

In optical experimental mechanics, the localization of points can be classified into monocular and multi-view measurements based on the number of cameras used. The most commonly used technique is triangulation. It involves forming a triangle with the optical centers of two cameras and the baseline, connecting two lines of sight, and the intersection of the two sight-rays is the position of the target point. As shown in Fig. \ref{fig1a}, when the point is static, triangulation can be achieved by moving the camera. In the triangulation algorithm, the intrinsic and extrinsic parameters of the camera need to be known. Therefore, camera calibration and self-calibration become important techniques for triangulation. These technologies also play crucial roles in many areas of photomechanics, such as structure from motion (SfM), visual odometry (VO), and simultaneous localization and mapping (SLAM). Currently, there are many mature camera calibration methods available. \cite{JW10,JW11,JW12}

However, as shown in Fig. \ref{fig1b}, synchronized observation from multiple cameras is required to achieve triangulation when the point moves. At this time, the dynamic 3D reconstruction problem is reduced to the case of static 3D reconstruction. Triangulation algorithms under such observation conditions have been systematically developed \cite{JW13,JW14,JW15,JW16}.

In tightly controlled laboratory setups, it is possible to have all cameras temporally synchronized. However, in practical applications, it isn't easy to synchronize cameras mounted on multiple UAVs. Thus, in most image sequences captured by multiple independent UAVs, no two cameras see the 3D point simultaneously. This fact trivially invalidates the triangulation constraint \cite{JW17}. Therefore, it is necessary to develop algorithms for localizing moving points based on asynchronous camera systems, which can significantly enlarge the applicability of UAVs.

The 3D trajectory reconstruction of moving points involves two coupled sub-problems: reconstructing the 3D trajectories of moving points and estimating the time information for each camera. Many previous works address one of these sub-problems. For instance, assuming the time information of the camera (frame rate and offset) is known, the linear or conic trajectory can be reconstructed using the monocular reconstruction algorithm proposed by Avidan and Shashua \cite{JW18}, called trajectory triangulation. Kaminski et al. \cite{JW19} extend this method to trajectories of arbitrary shapes. Moreover, methods that represent the motion trajectory of points using DCT trajectory basis vectors \cite{JW20,JW21,JW22} or temporal polynomials \cite{JW23,JW24,JW25} can be employed to reconstruct arbitrary trajectories. Regarding solving for camera time information, the most stable temporal alignment methods necessitate corresponding 2D trajectories as input, as indicated in Refs. \cite{JW26,JW27,JW28,JW29}. These methods depend solely on geometric cues to synchronize the interpolated points along the trajectories captured by different cameras. Albl et al. \cite{JW30} employ epipolar geometry as a constraint. However, this approach requires the camera to be stationary, which is not conducive to the application of UAVs. Zhou et al. \cite{JW31} propose to leverage a time-calibrated video featuring specific markers and a uniformly moving ball to accurately extract the temporal relationship between local and global time systems across cameras. This allows for the calculation of new timestamps and precise frame-level alignment. However, this method requires targets with special markers, which can also limit its range of application.

Li et al. \cite{JW32} propose an iterative method to address both sub-problems. They first assume that the multiple cameras are synchronized and use triangulation to reconstruct the trajectory. Given the known target trajectory, the time offset can be solved. Subsequently, with the known time offset, the target trajectory can be represented as a temporal polynomial for further solution. Through iteration, the final target trajectory and offset can be obtained. VO et al. \cite{JW33} propose a spatiotemporal bundle adjustment framework that optimizes camera parameters and target positions simultaneously. However, the motion priors they proposed do not apply to arbitrary motions. Moreover, it isn't easy to obtain a good initialization.

This paper proposes a method for simultaneously estimating camera time information and reconstructing target trajectories. By representing the target motion trajectory as temporal polynomials, the target trajectory can be reconstructed using several motion parameters. The reconstruction accuracy of the proposed method mainly depends on the rotation measurement accuracy of the optoelectronic platform. In typical application environments, UAVs often cannot carry high-precision camera rotation measurement devices, resulting in target positioning accuracy that frequently fails to meet application requirements. Previous work utilizes stationary points in the scene as constraints to optimize camera rotations \cite{JW34}. However, the moving points are often closer to the cameras, providing tighter constraints. Since the task is to locate moving points, multiple target points usually persist in multiple cameras' common field of view. Therefore, there can be sufficient geometric constraints. This paper introduces a novel framework that concurrently optimizes camera time parameters, rotation estimates, and target motion trajectories while leveraging geometric constraints inherent in image sequences. By integrating these elements, the approach enhances camera rotation accuracy, improving the precision of target localization through more reliable pose estimation. Both simulated and real-world experiments validate the feasibility and accuracy of the proposed method.

This paper proposes a trajectory reconstruction method for moving points based on asynchronous cameras. The main contributions of this paper include: 
\begin{enumerate}
    \item We extend the trajectory intersection method to asynchronous cameras, thereby achieving asynchronous trajectory reconstruction of moving points with known time information. This approach resolves the limitation of traditional triangulation that requires camera synchronization.
      
    \item We establish models for camera time information and target motion by analyzing imaging mechanisms and target dynamics. The parameters are simultaneously optimized, and the trajectory reconstruction without known time information is achieved with superior accuracy and efficiency compared to iterative methods.
      
    \item We propose a novel bundle adjustment framework that jointly optimizes camera rotations, time information, and target motion parameters, using tighter constraints on moving targets. The reconstruction accuracy is significantly improved, especially when the camera rotations are inaccurate.
\end{enumerate} 

The rest of this paper is arranged as follows. Section \ref{sec2} formulates the coupled sub-problems and proposes the method for simultaneously solving camera time information, rotations, and target motion parameters. Section \ref{sec3} validates the proposed method through simulated and real-world experiments, respectively. Section \ref{sec4} discusses the obtained results, analyzes the applicability of our method, and outlines future research directions. Ultimately, Section \ref{sec5} offers a conclusion.
\section{Theories and methods}\label{sec2}

This section formulates the problems to be solved and proposes the trajectory reconstruction method of moving points based on asynchronous cameras. Section \ref{sec2.1} formulates the coupled sub-problems to be solved in this paper. Section \ref{sec2.2} extends the trajectory intersection method from monocular to multi-camera to address trajectory reconstruction based on asynchronous cameras with known time information. Section \ref{sec2.3} models the camera time information and target motion, which are optimized simultaneously. Section \ref{sec2.4} proposes a method that simultaneously optimizes asynchronous cameras' time information, rotations, and targets' motion parameters.

\subsection{Problem Formulation}\label{sec2.1}
\begin{figure*}[t] 
\centering
\subfloat[UAVs synchronously observe the target]{\includegraphics[width=3in]{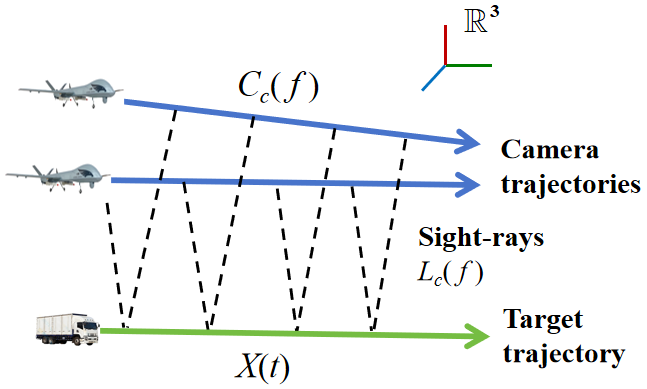}%
\label{fig2a}}
\hfil
\subfloat[UAVs asynchronously observe the target]{\includegraphics[width=3in]{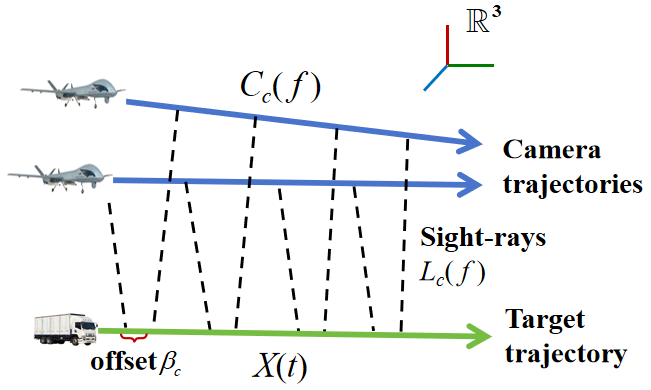}%
\label{fig2b}}
\caption{Illustration of 3D trajectory reconstruction of a moving target based on UAVs.}
\label{fig2}
\end{figure*}
\begin{figure}[H]
    \centering
  \includegraphics[width=2.5in]{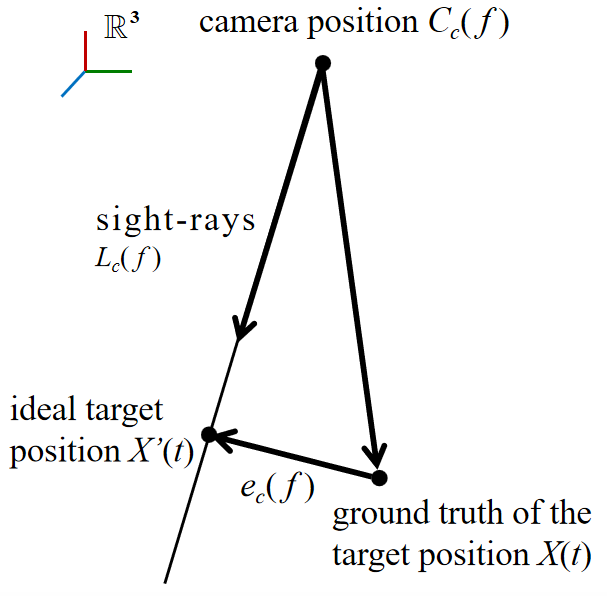}
\caption{Illustration of residual error between the ground truth and the ideal position of the target.}
\label{fig3}
\end{figure}

Figure \ref{fig2} illustrates the scenario where UAVs, equipped with cameras, are observing a moving target synchronously and asynchronously, respectively. $C_{c}(f)$ represents the camera trajectories, $X(t)$ represents the target trajectory, $L_{c}(f)$ represents the sight-rays, and $\beta_{c}$ represents the time offset of camera $c$. Consider the scenario of $C$ cameras observing the same 3D point over time. Camera $c$ captures an image of the target point at time $t$, constituting the $f-th$ frame in the image sequence of camera $c$. The relation between the 3D point $X(t)$ and its 2D projection $x_{c}(f)$ on camera $c$ at frame $f$ is given by:
\begin{equation}
\begin{bmatrix}x_{c}(f)\\1\end{bmatrix}\simeq K_{c}\begin{bmatrix}R_{c}(f)&T_{c}(f)\\1&1\end{bmatrix}\begin{bmatrix}X(t)\\1\end{bmatrix},
    \label{eq1}
\end{equation}
where $K_{c}$ represents the intrinsic matrix of camera $c$, which can be calibrated in advance. $R_{c}(f)$ and $T_{c}(f)$ represent the rotation matrix and translation vector of camera $c$ when it is capturing the $f-th$ frame image, which are also known as the extrinsic parameters. The global time $t_{c}(f)$ when the $f-th$ frame captured by camera $c$ can be represented as:
\begin{equation}
t_{c}(f)=f/\alpha_{c}+\beta_{c},
    \label{eq2}
\end{equation}
where $\alpha_{c}$ and $\beta_{c}$ are the camera frame rate and time offset, respectively. The camera frame rate $\alpha_{c}$ is usually a known quantity. In this research, the intrinsic parameters $K_{c}$ of each camera are obtained through prior calibration \cite{JW10,JW11,JW12,JW13}. The rotation matrix $R_{c}(f)$ and translation vector $T_{c}(f)$ of each camera can be calculated employing the technique of structure from motion within a given scene \cite{JW13,JW34} or obtained from the Global Navigation Satellite System (GNSS) and Inertial Measurement Unit (IMU). Therefore, we can calculate the position of each camera's optical center $C_{c}(f)$ and the direction of the observation sight-ray $L_{c}(f)$:
\begin{equation}
C_{c}(f)=-R_{c}^{-1}(f)T_{c}(f),
    \label{eq3}
\end{equation}
\begin{equation}
L_{c}(f)=\frac{R_{c}^{\mathrm{T}}(f)K_{c}^{-1}x_{c}(f)}{\parallel R_{c}^{\mathrm{T}}(f)K_{c}^{-1}x_{c}(f)\parallel}.
    \label{eq4}
\end{equation}
Figure \ref{fig3} illustrates the residual error between the ground truth and the ideal position of the target. $C_{c}(f)$ represents the camera position, $L_{c}(f)$ represents the sight-ray, $X(t)$ represents the ground truth of the target position, $X'(t)$ represents the ideal target position. $e_{c}(f)$ represents the residual error between the ground truth and the ideal position of the target. The residual error $e_{c}(f)$ can be represented as:
\begin{equation}
e_{c}(f)=(\mathrm{I}-L_{c}(f)L_{c}^{T}(f))(X(t)-C_{c}(f)),
    \label{eq5}
\end{equation}
$\mathrm{I}$ represents the 3x3 identity matrix. Under the criterion of minimizing the sum of squared residuals, we can establish the following equation to calculate the position of the target at each observation moment $X(t)$:
\begin{equation}
\mathrm{V}_{c}(f)X(t)=\mathrm{V}_{c}(f)C_{c}(f).
    \label{eq6}
\end{equation}
where $\mathrm{V}_{c}(f)=\mathrm{I}-L_{c}(f)L_{c}^{T}(f)$. When all cameras are synchronized, they capture images of the moving target simultaneously, as illustrated in Fig. \ref{fig2a}. At this point, assuming that each camera captures $F$ images of the target, the number of unknowns is $3F$. According to equation (6), each observation from each camera can establish 2 independent equations. Therefore, the total number of independent equations that can be established is $2FC$. Therefore, when $C > 1$, the least squares solution for the target's position $X(t)$ can be obtained through triangulation-based intersection.
\begin{figure*}[t] 
\centering
\subfloat[The camera motion is relatively simple]{\includegraphics[width=2.2in]{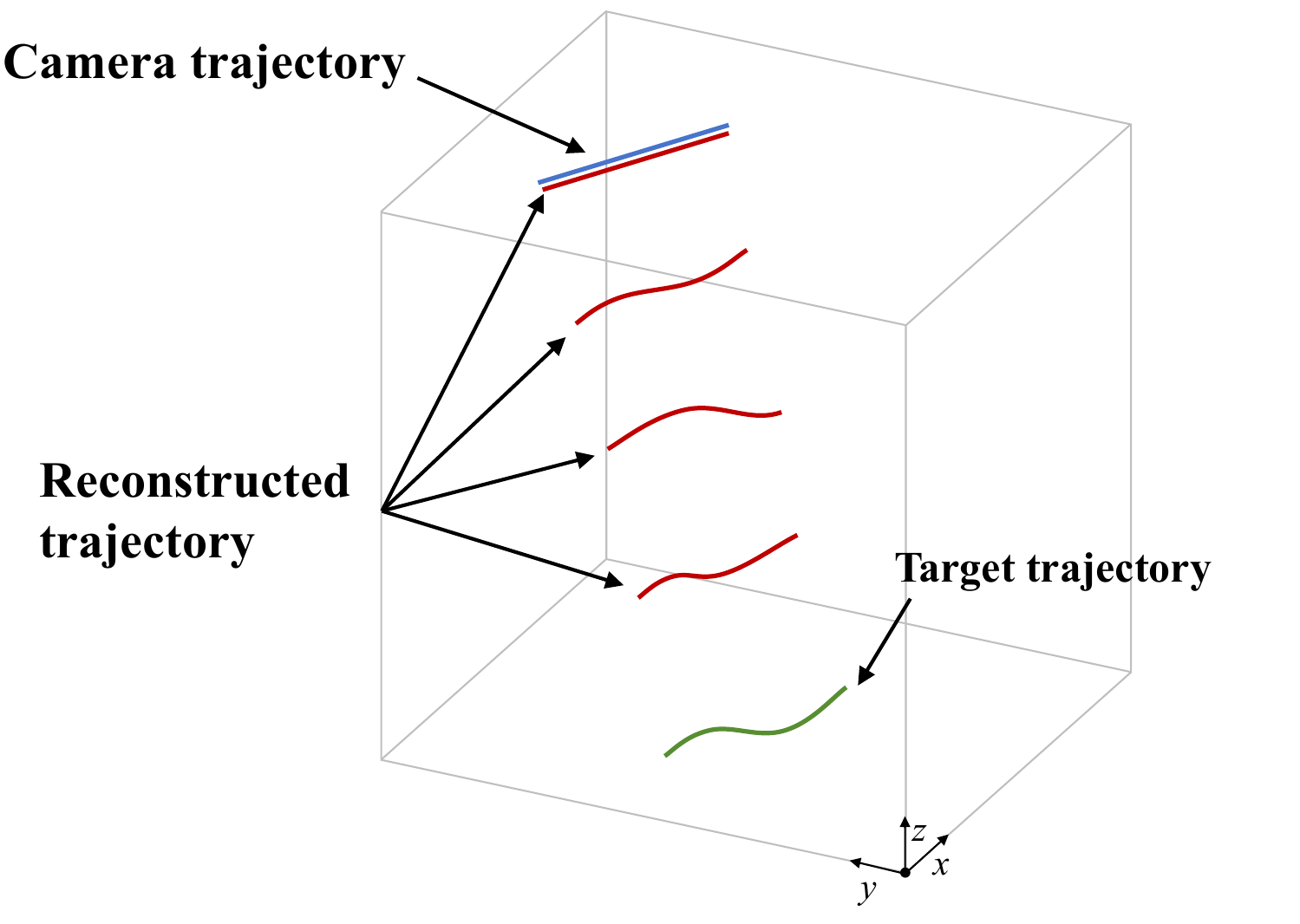}%
\label{fig4a}}
\hfil
\subfloat[All sight-rays intersect at the same point.]{\includegraphics[width=2.2in]{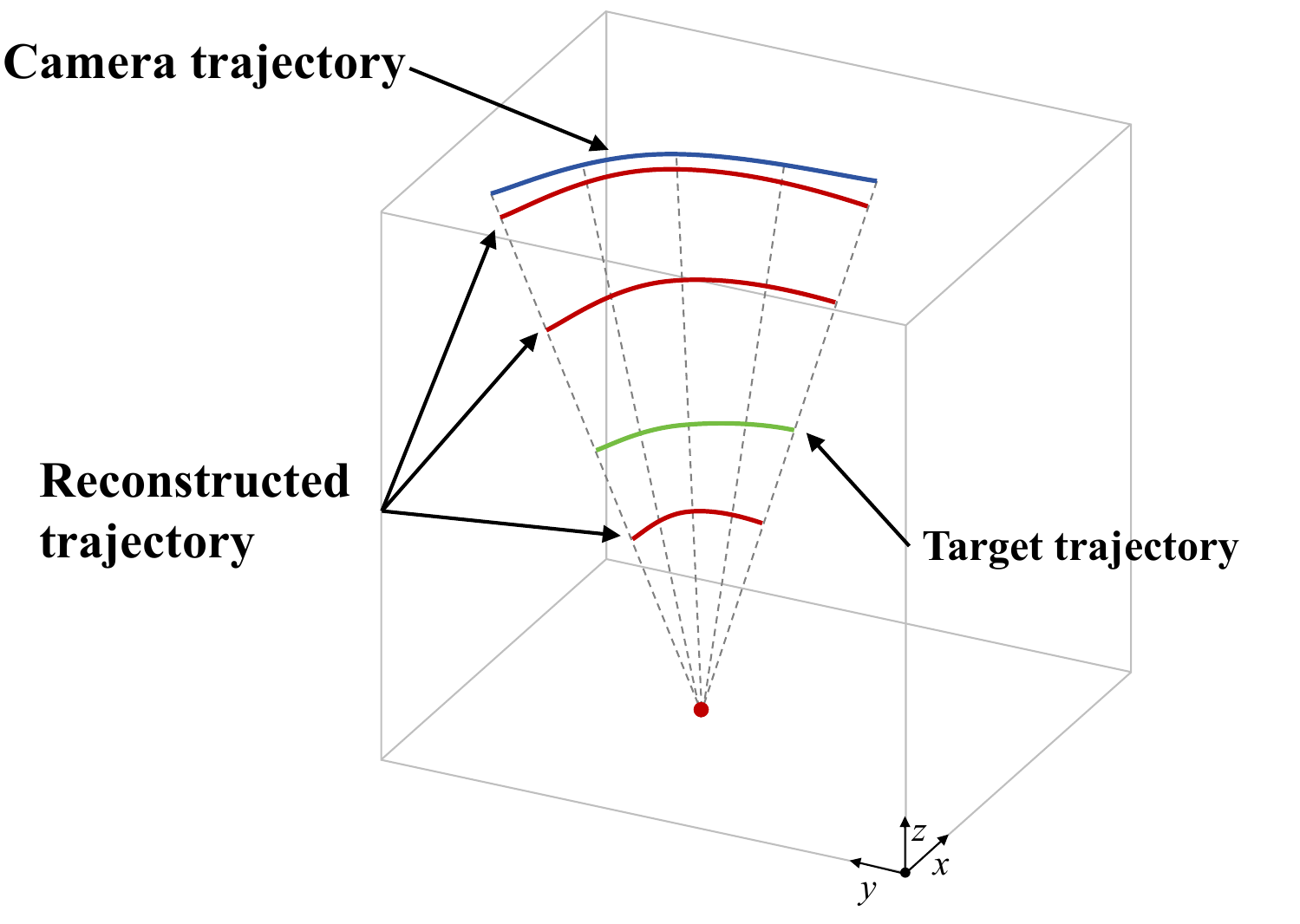}%
\label{fig4b}}
\hfil
\subfloat[All sight-rays are parallel]{\includegraphics[width=2.2in]{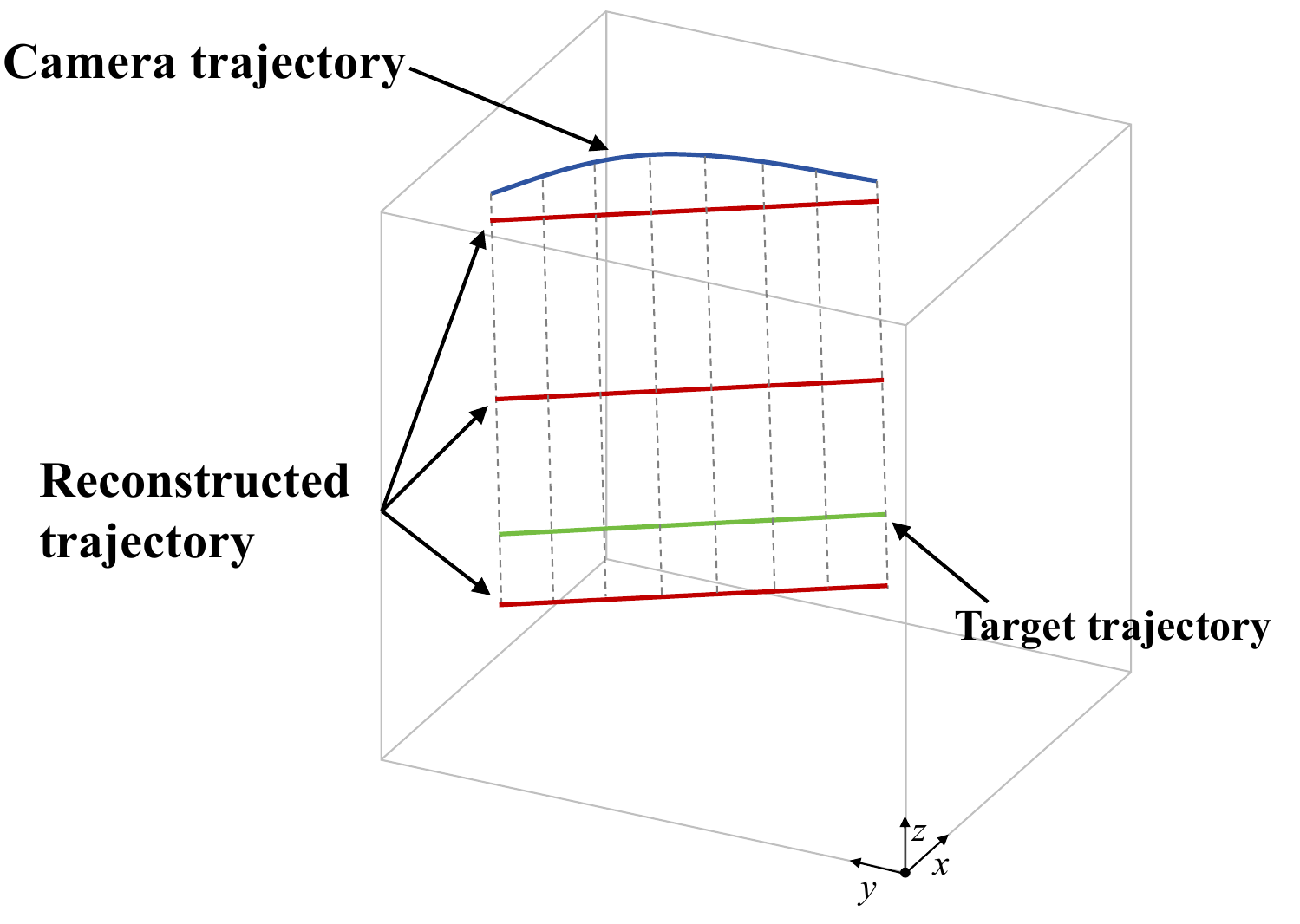}%
\label{fig4c}}
\caption{Degeneracy situations of monocular trajectory intersection.}
\label{fig4}
\end{figure*}
However, when all cameras are asynchronous, the traditional triangulation-based intersection constraints are no longer applicable, as shown in Fig. \ref{fig2b}. At this point, the number of unknowns is $3FC$, while the number of independent equations is only $2FC$. Thus, without reasonable assumptions about the target's motion, it is impossible to reconstruct the 3D trajectory of the target.

\subsection{Multi-camera trajectory intersection with known time information}\label{sec2.2}

When all cameras are asynchronous, i.e., the time offsets $\beta_{c}$ of each camera are not zero, the traditional synchronized triangulation constraints are no longer applicable. At this point, to solve for the target's trajectory $X(t)$, reasonable assumptions about the target's motion need to be made. Yu et al. \cite{JW23} propose the trajectory intersection method, which assumes that the target trajectory can be expressed as temporal polynomials to solve for the target's trajectory based on a monocular camera. The temporal polynomials' coefficients can be obtained by intersecting a series of sight-rays with the target's parameterized motion trajectory. The coefficients of the temporal polynomials have physical meanings, such as velocity, acceleration, etc. Since within a certain period of time, the ground-moving target's motion usually follows certain physical laws, such as static state, uniform linear motion, or uniform accelerated motion. Therefore, it is highly appropriate to represent the target's motion trajectory using temporal polynomials. 

When the time offsets $\beta_{c}$ of each camera are not zero but known, the image sequences captured by an asynchronous multi-camera system can be regarded as being captured by a monocular camera. The trajectory intersection method can be easily extended to a multi-camera system with known time information. The target's trajectory is represented by the following temporal polynomials:
\begin{equation}
\begin{cases}X_{i}=\sum\limits_{k=0}^{K}a_{k}t_{i}^{k}\\Y_{i}=\sum\limits_{k=0}^{K}b_{k}t_{i}^{k}\\Z_{i}=\sum\limits_{k=0}^{K}c_{k}t_{i}^{k}&\end{cases},
    \label{eq7}
\end{equation}
where $K$ is the order of the polynomials. $a_{k}(k=0,1,\cdots, K)$, $b_{k}(k=0,1,\cdots, K)$, $c_{k}(k=0,1,\cdots, K)$ are the motion parameters of the target that need to be solved. Based on Eq. (\ref{eq6}) and Eq. (\ref{eq7}), the subsequent set of equations can be formulated:
\begin{equation}
\mathrm{V}_{c}(f)\begin{bmatrix}\sum\limits_{k=0}^{K}a_{k}t_{i}^{k}\\\sum\limits_{k=0}^{K}b_{k}t_{i}^{k}\\\sum\limits_{k=0}^{K}c_{k}t_{i}^{k}\end{bmatrix}=\mathrm{V}_{c}(f)C_{c}(f).
    \label{eq8}
\end{equation}

Solving Eq. (\ref{eq8}) from $FC$ observations for the motion parameters of the target, the number of unknowns is 3($K$+1), while the number of independent equations is 2$NC$. It is a linear least squares system if $2FC \geq 3(K + 1)$. The estimated motion parameters $a_{k}$, $b_{k}$, and $c_{k}$ can be obtained by solving the linear least squares equations. The reconstructed trajectory of the target is a trajectory that passes through all sight-rays and is represented by linear polynomials of time $t_i$. Then the moving trajectory of the target can be reconstructed by Eq. (\ref{eq7}).

In the trajectory intersection method, the temporal polynomials are employed to represent the targets' motions. The order of the temporal polynomials $K$ is considered as a known quantity. Consistent with Ref. \cite{JW23}, the order $K$ of the temporal polynomials can be manually selected based on experience. For example, for targets such as vehicles on the road or ships on the surface of the sea, $K$ can be set to 1, i.e., the targets are assumed to be in uniform linear motion within a certain period of time. For targets with high maneuverability, $K$ can be set to 2, i.e., the uniform accelerated motion. This assumption holds reasonable validity over a short period of time.

It is worth mentioning that there are two situations where a definite solution cannot be obtained by the monocular trajectory intersection method, as shown in Figure 3:

(i) The order of the temporal polynomials representing the camera motion is equal to or lower than that of the target motion, as shown in Fig. \ref{fig4a}.

(ii) All sight-rays intersect at the same point (all sight-rays being parallel is a special case of this situation, where they intersect at the infinite point), as shown in Fig. \ref{fig4b} and \ref{fig4c}.

\begin{figure}[H]
    \centering
  \includegraphics[width=3.3in]{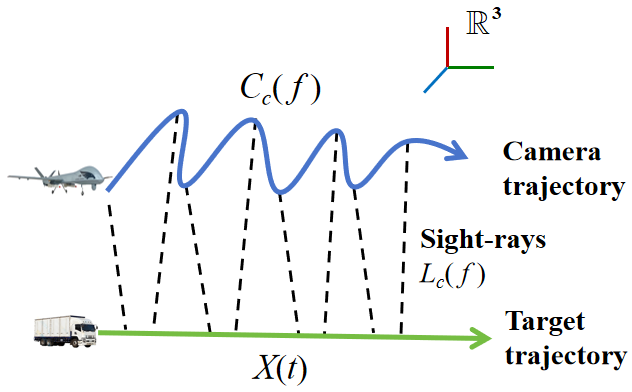}
\caption{Illustration of treating the asynchronous cameras as a monocular camera.}
\label{fig5}
\end{figure}

The degenerate cases of the multi-camera trajectory intersection with known time information are identical to those of the monocular camera method. When the cameras are asynchronous, only one camera observes the target at the same moment. We can treat the asynchronous multi-camera systems as a single monocular camera. As shown in Figure 5, the equivalent camera trajectory is a curve that connects various observation positions in chronological order. The motion trajectory of the equivalent monocular camera is complex, i.e., the equivalent order of the temporal polynomials is very high. Moreover, it is unlikely that all sight-rays intersect at the same point. Therefore, the above two degenerate cases are not likely to occur. Generally, using asynchronous multi-camera systems with known time information yields higher positioning accuracy than using only a monocular camera, as the motion of a single UAV has smoothness and continuity.

\subsection{Multi-camera trajectory intersection with unknown time information}\label{sec2.3}

Section \ref{sec2.2} extends the trajectory intersection method from a monocular camera to a multi-camera system, given that the time information of each camera is known. By approximating the short-term moving trajectory of a point target using temporal polynomials of different orders, it becomes possible to solve for the target's motion parameters and reconstruct its moving trajectory.

However, when the multi-camera system is asynchronous, each camera's time offsets $\beta_{c}$ are usually unknown. Therefore, the global timing of the images captured by each camera is unknown, which prevents the direct application of the multi-camera trajectory intersection method to reconstruct the target's trajectory. In this case, to reconstruct the target's motion trajectory, we need to estimate the global timing of each frame captured by every camera. 

Li et al. propose an iterative approach based on the multi-camera trajectory intersection method \cite{JW32}. They first assume that the multi-camera system is synchronized and use the multi-camera trajectory intersection method to compute the target's motion parameters. Subsequently, they treat the obtained target's motion parameters as known quantities and solve for the time offsets of each camera. By iteratively repeating the above two steps until convergence, they obtain the final target's motion parameters, which can then be used to reconstruct the target's trajectory through Eq. (\ref{eq7}). However, this method requires multiple iterations, with the second step involving a nonlinear solution. This paper proposes a bundle adjustment (BA) framework that simultaneously optimizes the target's motion parameters $a_{k}$, $b_{k}$, $c_{k}$, and the time offsets $\beta_{c}$. The objective function can be represented as follows:
\begin{equation}
S=\sum_{c=1}^{C}\sum_{f=1}^{F_{c}}\frac{1}{2}\|\mathrm{V}_{c}(f)(X_{c}(f)-C_{c}(f))\|^{2},
    \label{eq9}
\end{equation}
where $F_{c}$ represents the number of image frames captured by camera $c$. In the above objective function, the position of the target $X_{c}(f)$ at the moment when camera $c$ captures the $f-th$ frame can be represented as:
\begin{equation}
\begin{cases}X_{c}(f)=\sum\limits_{k=0}^{K}a_{k}t_{c}^{k}(f)\\Y_{c}(f)=\sum\limits_{k=0}^{K}b_{k}t_{c}^{k}(f)\\Z_{c}(f)=\sum\limits_{k=0}^{K}c_{k}t_{c}^{k}(f)&\end{cases},
    \label{eq10}
\end{equation}
where $a_{k}(k=0,1,\cdots, K)$, $b_{k}(k=0,1,\cdots, K)$, $c_{k}(k=0,1,\cdots, K)$ are the motion parameters of the target that need to be solved. The time parameter $t_{c}(f)$ can be calculated using the frame rate $\alpha_{c}$ and time offset $\beta_{c}$ of camera $c$ through Eq. (\ref{eq2}).

Therefore, the unknowns in the objective function (\ref{eq9}) are the target's motion parameters $a_{k}$, $b_{k}$, $c_{k}$ and the time information parameters $\alpha_{c}$, $\beta_{c}$ of each camera. We can solve for these parameters by minimizing the objective function (\ref{eq9}):
\begin{equation}
\arg\min_{a_{k},b_{k},c_{k},\alpha_{c},\beta_{c}}S
    \label{eq11}
\end{equation}
The intrinsic parameter $K_{c}$ is calibrated in advance. The rotation matrix $R_{c}(f)$ and translation vector $T_{c}(f)$ of the camera are calculated employing the technique of structure from motion within a given scene or obtained from the GNSS and IMU. The proposed BA framework requires many initial values, including the target's motion parameters $a_{k}$, $b_{k}$, $c_{k}$, and the cameras' time parameters $\alpha_{c}$ and $\beta_{c}$. The initial value of the frame rates $\alpha_{c}$ can be directly obtained from the camera. The initial value of the time offsets $\beta_{c}$ can be set to 0. At this point, we can use the multi-camera trajectory intersection method with known time information proposed in Section \ref{sec2.2} to calculate the initial values of target's motion parameters $a_{k}$, $b_{k}$, and $c_{k}$.

After obtaining the initial values, we can obtain the target's motion parameters $a_{k}$, $b_{k}$, $c_{k}$, and the time offsets $\beta_{c}$ of each camera by optimizing Eq. (\ref{eq11}) using the Levenberg-Marquardt algorithm. Then, the global time information can be calculated using Eq. (\ref{eq2}). The target's trajectory can be reconstructed using Eq. (\ref{eq10}). Compared with the method of Li et al., we do not require iterative calculations. Moreover, in cases where the camera frame rate $\alpha_{c}$ is inaccurate, we can optimize it together with other parameters. This can obtain a more precise frame rate, achieving a more accurate trajectory of the target's motion. It is worth mentioning that there is a degenerate case of the multi-camera trajectory intersection method when both the frame rate and the offset are unknown. When all the sight-rays are coplanar, a definite solution cannot be obtained. This is because any line on this plane satisfies the condition of zero residual. However, this situation can be easily avoided by controlling the motion of the flight platform.

\subsection{Multi-camera trajectory intersection with camera rotation optimization}\label{sec2.4}

The multi-camera trajectory intersection method proposed in Sect. \ref{sec2.3} is based on collinear equations. It utilizes the target dynamics characteristics over a short period of time and represents the target's motion as temporal polynomials. This approach optimizes the cameras' time information and the targets' motion parameters, thereby reconstructing the target's trajectory. Since one of the advantages of the trajectory intersection method is its ability to handle occlusions, the proposed method does not require the target point to always be within the common field of view of all cameras. 

\begin{figure}[H]
    \centering
  \includegraphics[width=3in]{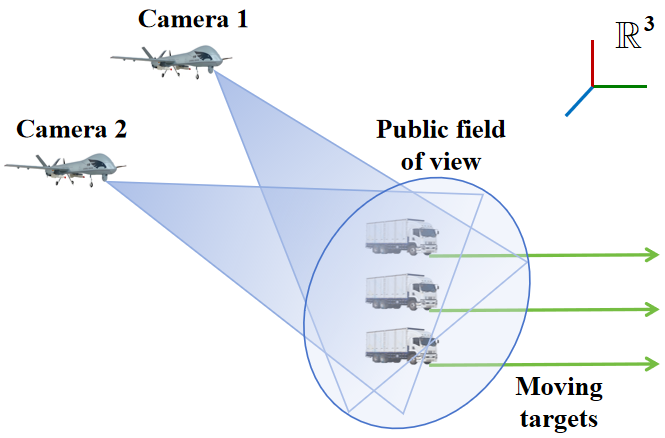}
\caption{Illustration of multi-camera system observing multiple targets.}
\label{fig6}
\end{figure}

The reconstruction accuracy of the proposed method mainly depends on the rotation measurement accuracy of the optoelectronic platform. Usually, the rotation of the platform is obtained from the IMU. In typical application environments, due to constraints such as size, weight, and power consumption, UAVs often cannot carry high-precision camera rotation measurement devices, resulting in target positioning accuracy that frequently fails to meet application requirements. Previous work utilizes stationary points in the scene as constraints to optimize camera rotations \cite{JW34}. However, finding feature points in the scene that can be continuously observed is difficult. The moving points are often closer to the cameras, providing tighter and more continuous constraints. As shown in Fig. \ref{fig6}, when a multi-camera system observes multiple targets, we have more continuous constraints available for use. We can utilize these constraints to enhance the precision of camera rotations, thereby improving the reconstruction accuracy.

This paper proposes a bundle adjustment framework that jointly optimizes the camera time information, rotations, and the target motion parameters. Make full use of the constraints in the image sequence to improve the accuracy of the camera rotations, thereby enhancing the localization accuracy of the targets. In typical application scenarios, the targets to be localized are usually multiple moving objects, such as vehicle convoys, fleets, etc. When cameras observe multiple moving targets, the geometric constraints formed by the precise camera positions provided by the GNSS and the multiple views of the multiple targets can be fully utilized. While solving for the target motion parameters, the camera time information and rotations can be optimized simultaneously as optimization parameters, thereby further improving the target localization accuracy.

Consider there are $C$ cameras observing $N$ moving points. The motion trajectories of each point target are represented using temporal polynomials of different orders as follows:
\begin{equation}
\begin{cases}X_{ni}=\sum\limits_{k=0}^{K_n}a_{nk}t_{i}^{k}\\Y_{ni}=\sum\limits_{k=0}^{K_n}b_{nk}t_{i}^{k}\\Z_{ni}=\sum\limits_{k=0}^{K_n}c_{nk}t_{i}^{k}&\end{cases},
    \label{eq12}
\end{equation}
where $(X_{ni},Y_{ni},Z_{ni})(n=1,2,\cdots,N)$ denotes the trajectory of the $n-th$ target. $a_{nk}(k=0,1,\cdots,K_n)$, $b_{nk}(k=0,1,\cdots,K_n)$, and $c_{nk}(k=0,1,\cdots,K_n)$ represent the coefficients of the temporal polynomials for the $n-th$ target, i.e., the motion parameters. $K_n$ represents the order of temporal polynomials for the $n-th$ target. Quaternions can avoid the pitfalls of singularity and are suitable for application scenarios that require high precision, high efficiency, and high stability \cite{JW35,JW36}. Thus, we represent the rotations of the cameras using quaternions. Let the rotations of camera $c$ when capturing the $f-th$ frame of the image sequence be represented by a quaternion as $q_{c}(f)=w+xi+yj+zk$. The conversion relationship between a quaternion and a rotation matrix is as follows:
\begin{equation}
R_{c}(f)=\begin{bmatrix}1-2y^{2}-2z^{2}&2xy-2wz&2xz+2wy\\2xy+2wz&1-2x^{2}-2z^{2}&2yz-2wx\\2xz-2wy&2yz+2wx&1-2x^{2}-2y^{2}\end{bmatrix}.
    \label{eq13}
\end{equation}

Since the localization accuracy of the targets primarily depends on the rotation accuracy of the cameras, to enhance the rotation accuracy, we utilize the geometric constraints of multiple cameras and multiple targets to simultaneously optimize the time information and rotations of each camera, as well as the motion parameters of each target. The objective function can be represented as follows:
\begin{equation}
S^{*}=\sum_{c=1}^{C}\sum_{n=1}^{N}\sum_{f=1}^{F_{c}}\frac{1}{2}\|\mathrm{V}_{c}^{n}(f)(X_{c}^{n}(f)-C_{c}(f))\|^{2},
    \label{eq14}
\end{equation}
where $\mathrm{V}_{c}^{n}(f)=\mathrm{I}-L_{c}^{n}(f)L_{c}^{n}(f)$. By minimizing Eq. (\ref{eq14}), the optimized time information and rotations of each camera, as well as the motion parameters of each target, can be obtained:
\begin{equation}
\arg\min_{q_{c}(f),a_{nk},b_{nk},c_{nk},\alpha_{c},\beta_{c}}S^{*}
    \label{eq15}
\end{equation}

In the above framework, each camera can establish two constraint equations for each observation of a target. Therefore, the number of independent constraint equations is $2\sum\limits_{c=1}^{C}F_{c}N$. Based on the physical laws governing the moving targets, we assume that temporal polynomials of a certain order can describe the trajectories of each target over a period of time. Therefore, the number of motion parameters to be solved for the target is $\sum\limits_{n=1}^{N}3(K_{n}+1)$. Since the camera rotation $q_{c}(f)$ has 3 degrees of freedom, the number of camera rotation parameters to be optimized is $3\sum\limits_{c=1}^{C}F_{c}$. The number of time information parameters for the camera is $2C$. In summary, the total number of parameters to be optimized is $\sum\limits_{n=1}^{N}3(K_{n}+1)+3\sum\limits_{c=1}^{C}F_{c}+C$. 

Next, we discuss the solvability of the framework above. When the number of targets $N=1$, the number of independent equations is $2\sum\limits_{c=1}^{C}F_{c}$, which is unsolvable. When the number of targets $N>1$, the framework above is solvable when the sum of the number of image frames captured by each camera satisfies $\sum\limits_{c=1}^{C}F_{c}\geq\frac{\sum\limits_{n=1}^{N}3(K_{n}+1)+2C}{2N-3}$. As the number of captured images increases, the number of constraint equations increases, improving solution accuracy.

In addition, since the trajectory intersection method can handle occlusions, the proposed algorithm is highly flexible. Not all targets need to be present in the field of view of each camera at all times. We only need to formulate Eq. (\ref{eq14}) for the targets that appear in the field of view of any camera. When the total number of equations satisfies the solvability conditions, we can utilize the motion parameters of the targets to reconstruct the complete trajectories.
\section{Experimental verification and results analysis}\label{sec3}

In order to demonstrate the robustness and accuracy of our methods, we conduct a series of experiments, including simulated and real-world experiments. In this section, the proposed algorithm without rotation optimization in Section \ref{sec2.3} is referred to as Algorithm 1, and the proposed algorithm with rotation optimization in Section \ref{sec2.4} is referred to as Algorithm 2. Sections \ref{sec3.1} and \ref{sec3.2} provide detailed analyses of the simulation and real-world experiment results, respectively.

\subsection{Simulated experiments}\label{sec3.1}
\begin{figure}[H]
    \centering
  \includegraphics[width=3in]{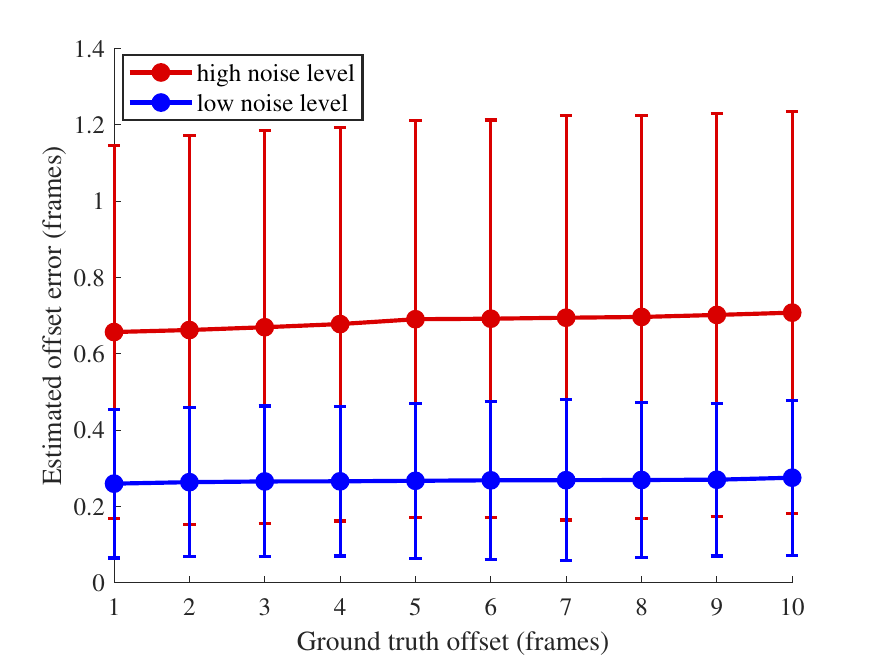}
\caption{Evaluation of the offset estimation.}
\label{fig7}
\end{figure}

In this section, we evaluate the performance of the proposed method on simulated data. This paper proposes a BA approach that optimizes the cameras' time information, rotations, and target motion parameters. We first evaluate the performance of camera time information estimation. We take the observation of a single moving target using asynchronous binocular cameras as an example. We introduce a variety of noises satisfying the normal distribution with a mean of zero to the simulated observation data, including target extraction noise, systematic noise and random noise in camera rotations, as well as systematic noise and random noise in camera position. The camera frame rate is set to 10 Hz, with an observation duration of 5 seconds, meaning each camera captures 50 frames of images. Without loss of generality, we set the offset of Camera 1 to 0, and the offsets of Camera 2 are set to 1-10 frames, respectively. Simulations are conducted under high noise levels and low noise levels, respectively. We use the algorithm proposed in Sect. \ref{sec2.3} to optimize the offsets. We conduct 1000 independent simulated experiments under high noise levels and low noise levels, respectively. The estimation results of the offsets are shown in Fig. \ref{fig7}. The red and blue curves represent the mean values of the estimated offset errors under high and low noise levels, respectively. The error bars represent the corresponding standard deviations.

As shown in Fig. \ref{fig7}, the estimated offset error increases with the growth of the ground truth offset and the noise level. Both the mean value and standard deviation of the estimated offset error is sensitive to observation noise but is less affected by the cameras' ground truth offsets. We can estimate the offset with relatively high accuracy in situations with low noise levels and a large ground truth offset. This may be due to the algorithm converging to a joint optimum of time offset and target motion parameters, rather than a global optimum of the time offset. The results demonstrate that the proposed algorithm can effectively estimate the cameras' offset and significantly reduce the impact of offset on reconstruction error.

\begin{figure*}[t] 
\centering
\subfloat[The result of uniform line motion target]{\includegraphics[width=3in]{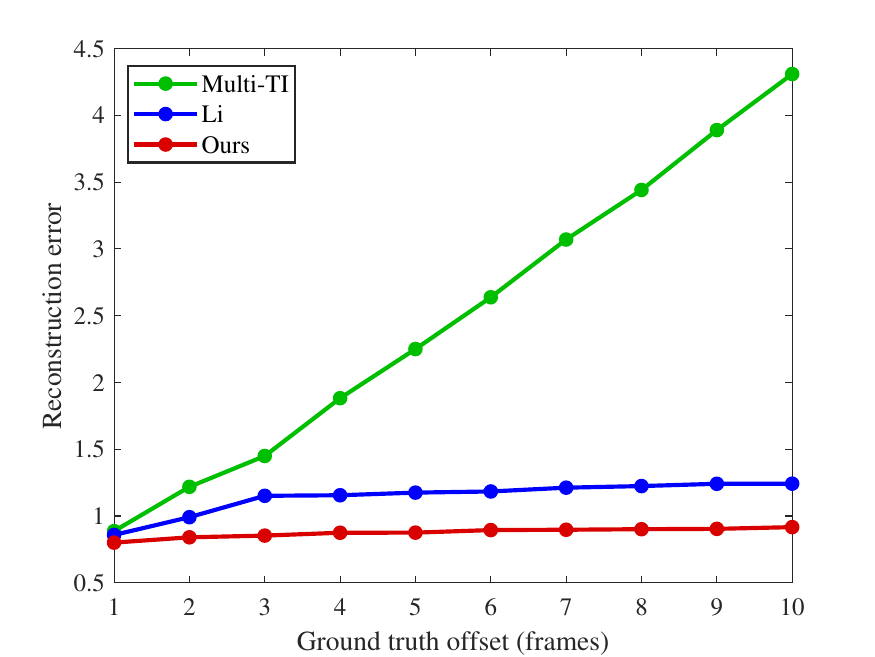}%
\label{fig8a}}
\hfil
\subfloat[The result of uniform acceleration motion target]{\includegraphics[width=3in]{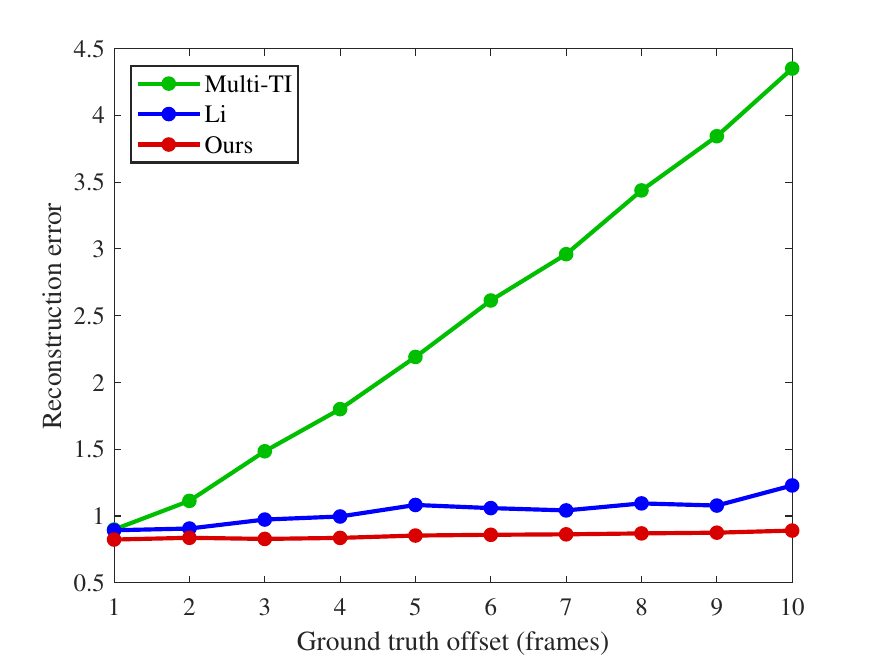}%
\label{fig8b}}
\caption{Evaluation of the reconstruction accuracy for varying motion across different offsets.}
\label{fig8}
\end{figure*}

\begin{figure*}[t] 
\centering
\subfloat[The result of uniform line motion target]{\includegraphics[width=3in]{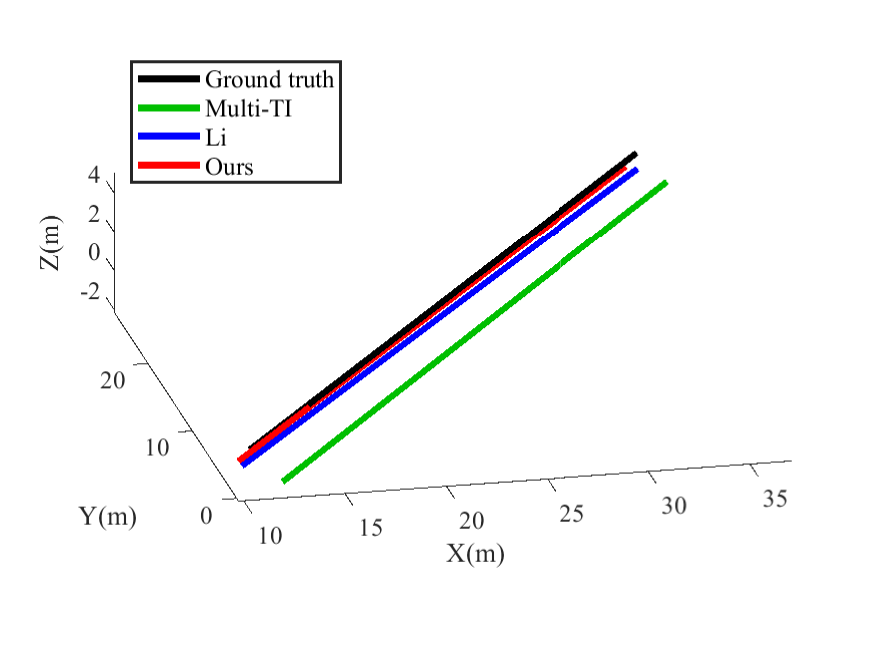}%
\label{fig9a}}
\hfil
\subfloat[The result of uniform acceleration motion target]{\includegraphics[width=3in]{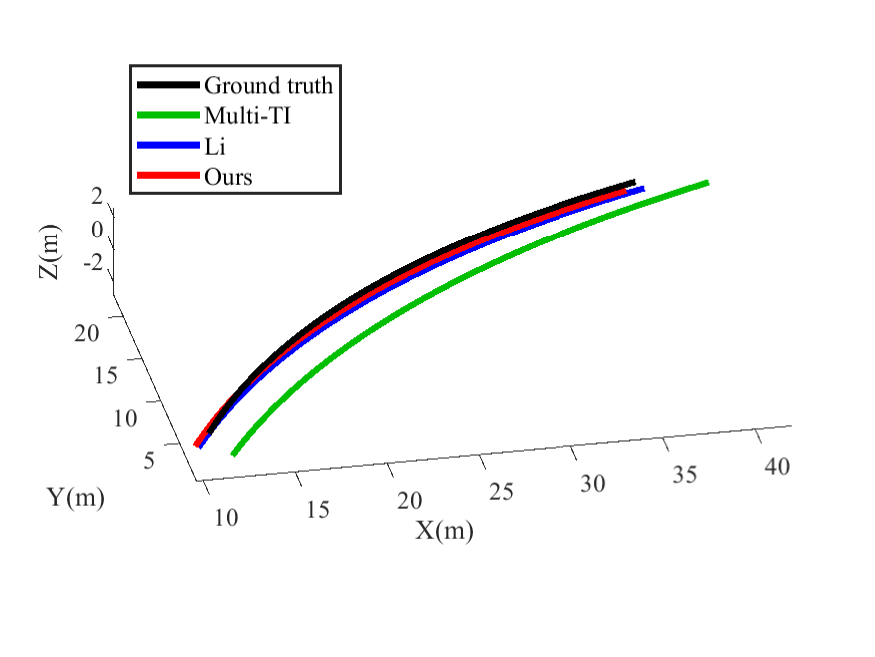}%
\label{fig9b}}
\caption{Illustration of the reconstruction results.}
\label{fig9}
\end{figure*}
To further validate the proposed offset estimation algorithm, we utilize it to simultaneously estimate the camera time offset and the target motion parameters, thereby reconstructing the target motion trajectory. As a comparison, we assume that the two cameras are synchronized and employ the multi-view trajectory intersection method (referred to as Multi-TI) introduced in Sect. \ref{sec2.2}, and Li et al.'s iterative solution method \cite{JW32} (referred to as Li) to reconstruct the target's motion trajectory. We use the mean of the localization errors of all viewpoints on the trajectory as the reconstruction error for that trajectory segment. Variety noises satisfying the normal distribution with a mean of zero are introduced to the simulated observation data, including target extraction noise with a standard deviation of 2 pixels, systematic noise and random noise in camera rotations with a standard deviation of 0.5°, systematic noise in camera position with a standard deviation of 3 m, as well as random noise in camera position with a standard deviation of 1 m. For different ground truths of offsets, we conduct 1,000 experiments, and calculate the average localization errors for the Multi-TI, Li, and proposed methods. The results are shown in Fig. \ref{fig8}, where Fig. \ref{fig8a} presents the reconstruction results for a target moving with uniform linear motion ($K$=1), and Fig. \ref{fig8b} presents the reconstruction results for a target moving with uniform acceleration ($K$=2). Figure \ref{fig9} illustrates the ground truth of the target trajectory and the reconstructed trajectories using the three methods.

\begin{figure*}[t] 
\centering
\subfloat[The result of uniform line motion target]{\includegraphics[width=3in]{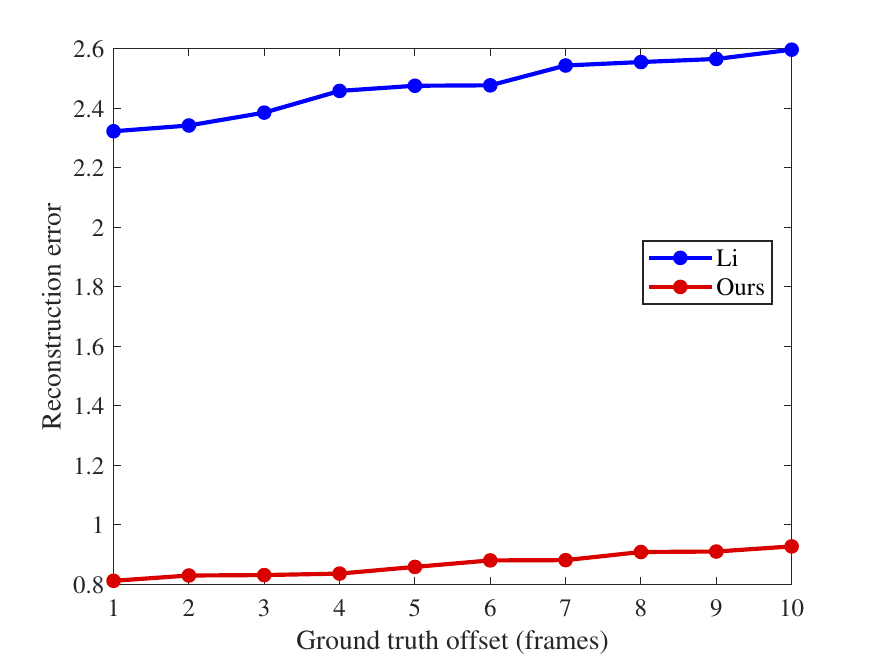}%
\label{fig10a}}
\hfil
\subfloat[The result of uniform acceleration motion target]{\includegraphics[width=3in]{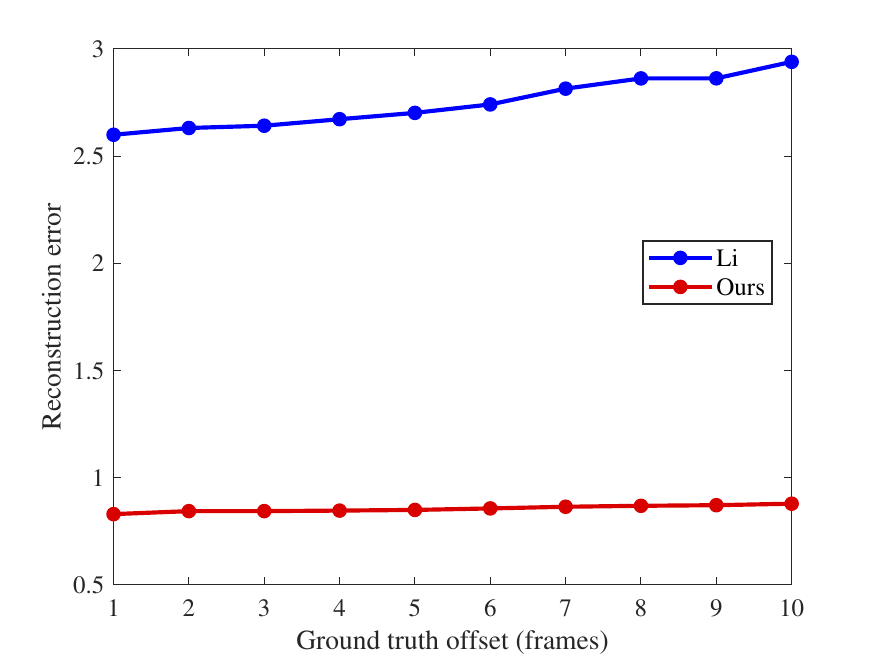}%
\label{fig10b}}
\caption{Evaluation of the reconstruction accuracy for varying motion across different offsets under conditions of inaccurate frame rate.}
\label{fig10}
\end{figure*}

\begin{figure*}[t] 
\centering
\subfloat[The result of uniform line motion target]{\includegraphics[width=3in]{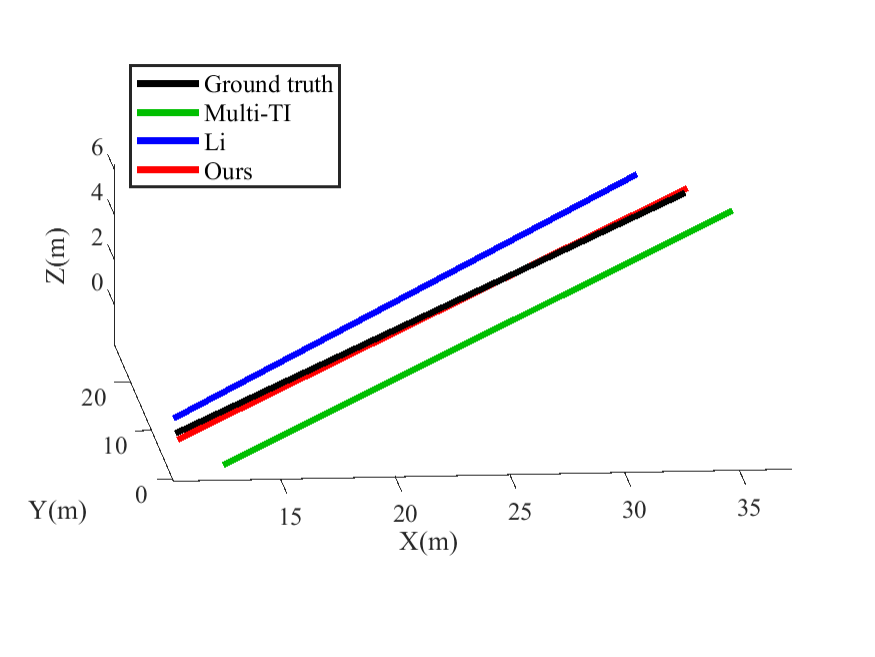}%
\label{fig11a}}
\hfil
\subfloat[The result of uniform acceleration motion target]{\includegraphics[width=3in]{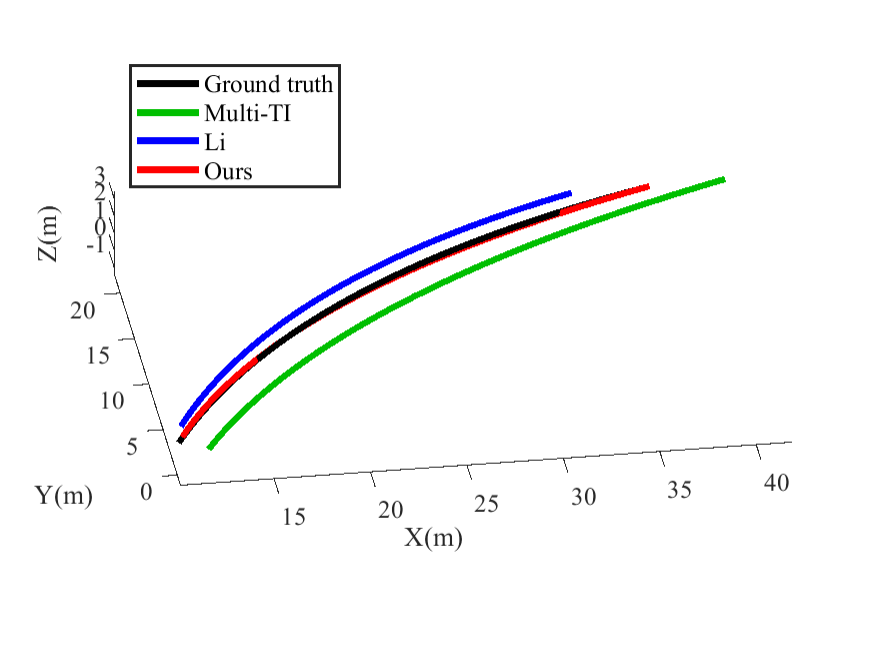}%
\label{fig11b}}
\caption{Illustration of the reconstruction results under conditions of inaccurate frame rate.}
\label{fig11}
\end{figure*}

As shown in Fig. \ref{fig8}, the proposed method achieves the highest reconstruction accuracy. The reconstruction error of the Multi-TI algorithm increases sharply as the offset increases. The accuracy of the proposed algorithm is less affected by the offset. This is consistent with Fig. \ref{fig7} that the offset estimation error of the proposed algorithm is insensitive to the ground truth offset. The reconstruction error of the Li method is also insensitive to variations in the ground truth offset. However, our algorithm consistently achieves higher reconstruction accuracy. As shown in Fig. \ref{fig9}, it can also be visually observed from the trajectory plot that the proposed method achieves the highest reconstruction accuracy. 

In addition, the proposed algorithm can optimize the cameras' frame rate. Therefore, our algorithm can achieve higher reconstruction accuracy when the frame rate is inaccurate. To evaluate the performance of the proposed algorithm under conditions of inaccurate frame rate, we compare it with the Li method. We still set the ground truth frame rate to 10 Hz, while the input to the algorithm is set to 9 Hz. The reconstruction errors of the two algorithms are shown in Fig. \ref{fig10}, and the reconstructed target trajectories are shown in Fig. \ref{fig11}.

As shown in Fig. \ref{fig10}, when the frame rate is inaccurate, the accuracy of the Li algorithm significantly decreases. However, the proposed algorithm treats the frame rate as an optimization parameter. Therefore, it demonstrates high robustness to frame rate accuracy. In situations where the frame rate is inaccurate, the reconstruction accuracy of the proposed algorithm is minimally affected. As shown in Fig. \ref{fig11}, under conditions of inaccurate frame rate, the trajectories estimated by the Multi-TI algorithm and the Li algorithm significantly deviate from the ground truth. However, the trajectory reconstructed by our algorithm is very close to the ground truth.

\begin{table}[H]
    \centering
    \caption{Evaluation of the calculation speed.}
    \label{tab1}
    \begin{tabular}{ccc}
        \toprule
      Motion Order & 1 & 2  \\
      \midrule
       Li (s)  & 0.0543 & 0.0771\\
       Ours (s)  & 0.0128 & 0.0186 \\
    \bottomrule 
    \end{tabular}
\end{table}

In addition, our algorithm exhibits higher computational efficiency than that of the Li method. To verify the computational efficiency of our algorithm, we conduct simulation experiments on a 12th Gen Intel(R) Core(TM) i9-12900H CPU. We run the Li algorithm and our algorithm separately, conducting 1,000 independent experiments, and calculating each algorithm's average runtime. The results are shown in Table \ref{tab1}. 

As shown in Table \ref{tab1}, the average runtime of the Li algorithm is more than twice that of our algorithm. This is because our algorithm needs to simultaneously optimize both the target motion parameters and the camera time information, but the Li algorithm requires multiple iterations. In simulation experiments, the Li algorithm typically requires 4-8 iterations. In addition, in the Li algorithm, calculating time offset also requires solving a system of nonlinear equations. In contrast, the BA framework we propose does not require iterative computation. Therefore, our algorithm demonstrates higher computational efficiency.

The above experimental results validate the high accuracy and robustness of the proposed algorithm under conditions such as asynchronous cameras and inaccurate frame rates. However, the accuracy of target trajectory reconstruction also depends on the precision of camera rotations, which are usually obtained from the IMU. In typical application environments, due to constraints such as size, weight, and power consumption, UAVs often cannot carry high-precision camera rotation measurement devices, resulting in target positioning accuracy that frequently fails to meet application requirements. When a multi-camera system observes multiple targets, we have more geometric constraints available for use. We can make full use of the geometric constraints in the image sequence to improve the accuracy of the camera rotations, thereby enhancing the localization accuracy of the targets.

This paper proposes a framework that optimizes the cameras' time information, rotations, and target motion parameters. We leverage the geometric constraints from multi-camera observations of multiple targets to optimize the inaccurate camera rotations, further enhancing the accuracy of trajectory reconstruction. To evaluate the performance of the proposed method, we compare the Li algorithm with our Algorithm 1 and Algorithm 2. We simulate a scenario where two asynchronous cameras observe four moving targets. The above variety noises satisfying the normal distribution with a mean of zero are introduced to the simulated observation data. The trajectories of the moving targets are reconstructed using the three methods above. To quantitatively compare the reconstruction accuracy, we select the average localization error of viewpoints along the trajectory of one of the targets as the metric for reconstruction error. Conduct 1000 independent experiments and calculate the average reconstruction error for each method. The experimental results are presented in Table \ref{tab2}.

\begin{table}[H]  
\centering  
\caption{Evaluation of the reconstruction accuracy under inaccurate camera rotations.} 
\label{tab2}  
\begin{tabular}{ccccc}  
\toprule
Motion Order & \multicolumn{2}{c}{1} & \multicolumn{2}{c}{2} \\ 
    \midrule
Accurate FPS? & yes & no & yes & no \\  
Multi-TI & 4.5603 & 6.4130 & 5.8066 & 7.1638 \\  
Li & 3.7426 & 4.8395 & 4.9393 & 5.8475 \\  
Algorithm 1 & 3.1392 & 3.9755 & 4.3099 & 4.6972 \\  
Algorithm 2 & 1.0085 & 1.1759 & 1.5983 & 1.6865 \\  
    \bottomrule 
\end{tabular} 
\end{table}  

As shown in Table \ref{tab2}, consistent with the previous conclusions, our Algorithm 1 exhibits higher accuracy than the Li algorithm, especially when the frame rate is inaccurate. This is because our algorithm incorporates the frame rate as a parameter for optimization, thereby enhancing the accuracy of the frame rate. Moreover, our algorithm further improves the reconstruction accuracy after optimizing the camera rotations. This demonstrates that our algorithm fully leverages the additional geometric constraints from multiple targets, effectively enhancing the accuracy of camera rotations, thereby improving the reconstruction accuracy of target trajectories.
\begin{figure*}[t]
    \centering
  \includegraphics[width=5.5in]{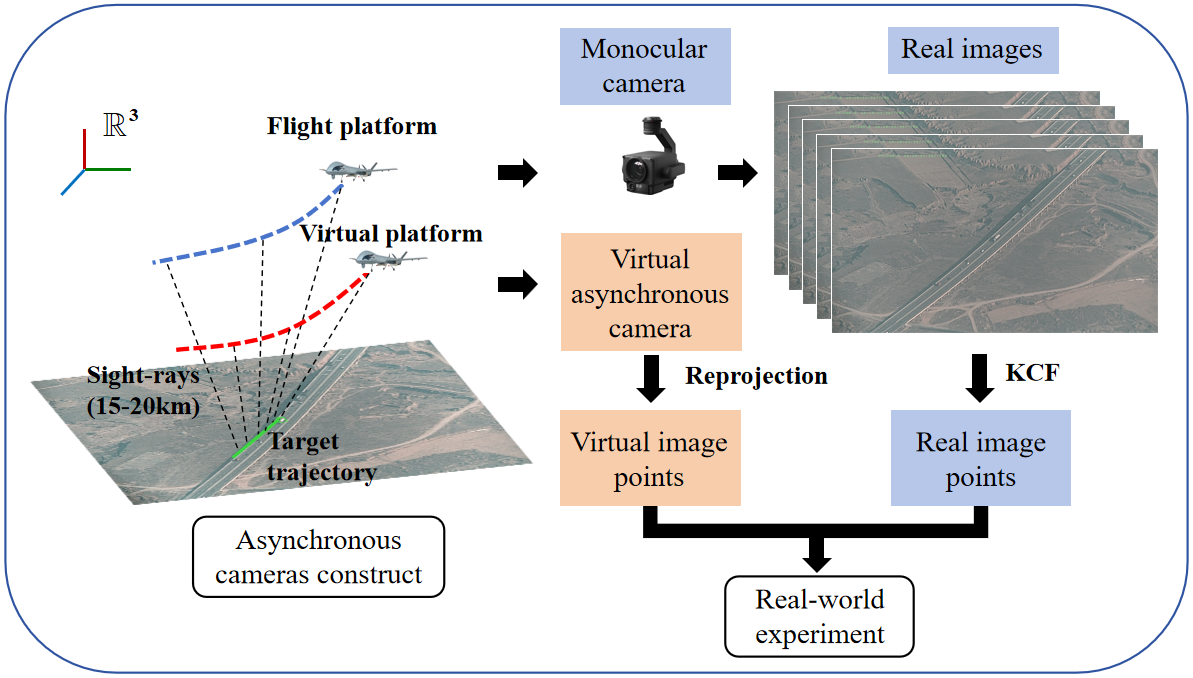}
\caption{Illustration of constructing the real-world experimental data.}
\label{fig12}
\end{figure*}

\subsection{Real-world experiments}\label{sec3.2}

In this section, we evaluate our method on real-world data. In the real-world data, the UAV observation platform observes vehicles traveling on the highway. Our experimental setup is illustrated in Fig. \ref{fig12}. The flight platform is equipped with a monocular RGB camera, which captures images at a spatial resolution of 1280 $\times$ 720 pixels and a temporal resolution of 25 Hz. Due to the difficulty in determining the ground truth offset of asynchronous cameras, we construct a binocular camera system consisting of a virtual camera and the monocular camera mounted on the platform, as shown in Fig. \ref{fig12}.

As shown in Fig. \ref{fig12}, the virtual camera has a spatial resolution of 1280$\times$720 and a temporal resolution of 25 Hz. We set the binocular camera system to operate asynchronously with a time offset of 40 frames. The flight platform is also equipped with GNSS and IMU to provide the monocular camera with pose information. The moving target travels along the highway in approximately uniform straight-line motion, with a speed of approximately 60 km/h. The flight platform performs curved motion while observing the moving target. The motion trajectory of the virtual platform shares the same shape as that of the flight platform, with a baseline length of 20 km between them. The observation distance range of the platforms extends from 15 to 20 km. To test the performance of Algorithm 2, three virtual targets are set up around the actual target. These virtual targets move at a constant speed in a straight line, with velocities of approximately 60 km/h.

The position of the camera's optical center and the ground truth of the target trajectory are provided by GNSS. The Kernelized Correlation Filters (KCF) algorithm \cite{JW37} is used to track the target. Under such observation conditions, the target on the road can be regarded as a point target. We are only concerned with the trajectory and motion parameters of the target, regardless of its rotation. The center of the bounding box is taken as the image point of the target. The image points coordinate of the target points on the virtual camera are obtained through reprojection, and noise is introduced. We use image sequences of target points on real and virtual cameras to reconstruct the target's trajectory. We reconstruct the target trajectory using the Multi-TI algorithm, our Algorithm 1 and Algorithm 2. The reconstruction errors of each method are calculated separately, and the results are shown in Table \ref{tab3}. $\sigma_x(m)$, $\sigma_y(m)$, $\sigma_z(m)$ represent the localization errors in the x, y, and z directions, respectively. $\sigma(m)$ represents the 3D localization error.
\begin{table}[H]  
\centering  
\caption{Evaluating the reconstruction accuracy in real-world experiments.}  
\label{tab3}  
\begin{tabular}{ccccc}  
\toprule
Method & $\sigma_x(m)$ & $\sigma_y(m)$ & $\sigma_z(m)$ & $\sigma(m)$\\ 
    \midrule
Multi-TI & 210.23 & 28.93 & 128.62 & 248.41 \\  
Algorithm 1 & 153.36 & 68.29 & 68.29 & 217.12 \\  
Algorithm 2 & 90.04 & 34.13 & 58.10 & $\mathbf{112.95}$ \\  
    \bottomrule 
\end{tabular} 
\end{table}  
It can be seen in Table \ref{tab3} that the reconstruction accuracy of the Multi-TI algorithm is the lowest. This may be because we construct asynchronous cameras. However, the Multi-TI algorithm assumes synchronization between the virtual and real cameras, which results in significant reconstruction errors. Our Algorithm 1 models the time information of asynchronous cameras. Then, treat the time parameters as optimization parameters, simultaneously optimizing them alongside the target motion parameters. This algorithm effectively reduces the reconstruction error caused by camera asynchrony. Therefore, the reconstruction error of Algorithm 1 is lower than that of the Multi-TI algorithm. 
\begin{table}[H]  
\centering  
\caption{Evaluating the reconstruction accuracy in real-world experiments under conditions of inaccurate frame rate.}  
\label{tab4}  
\begin{tabular}{ccccc}  
\toprule
Method & $\sigma_x(m)$ & $\sigma_y(m)$ & $\sigma_z(m)$ & $\sigma(m)$\\ 
    \midrule
Multi-TI & 239.65 & 97.46 & 188.61 & 320.16 \\  
Algorithm 1 & 193.88 & 77.07 & 139.83 & 251.16 \\  
Algorithm 2 & 104.18 & 48.41 & 113.21 & $\mathbf{161.29}$ \\  
    \bottomrule 
\end{tabular} 
\end{table}  
In typical application environments, due to constraints such as size, weight, and power consumption, UAVs often cannot carry high-precision camera rotation measurement devices. Therefore, the rotation errors of the cameras tend to be relatively large, which can lead to significant reconstruction errors. Our algorithm with rotation optimization fully utilizes the constraints of multiple targets, and treats camera rotations as optimization parameters. It simultaneously optimizes the camera time information, rotations, and target motion parameters. As shown in Table \ref{tab3}, when the camera rotations are inaccurate, the proposed method optimizes the camera rotations, thereby significantly improving the reconstruction accuracy. In the experimental scenario, with an observation distance of 15 $\sim$ 20 km, the localization error is 112.95 m. To evaluate the performance of the proposed method under conditions of inaccurate frame rate, we set the input frame rate to 20 Hz. The experimental results under conditions of inaccurate frame rate are presented in Table \ref{tab4}.

As shown in Tables \ref{tab3} and \ref{tab4}, due to the influence of inaccurate frame rate conditions, the reconstruction accuracy of all three algorithms decreases. The condition of inaccurate frame rate has the most significant impact on the Multi-TI algorithm. Since we incorporate the frame rate as a parameter for optimization, our Algorithm 1 and Algorithm 2 are less affected by conditions of inaccurate frame rate. In addition, since Algorithm 2 optimizes camera rotation by leveraging constraints on motion points, its accuracy remains significantly higher than that of the Multi-TI algorithm and Algorithm 1. The real-world experimental results demonstrate the effectiveness and high accuracy of the proposed algorithm, especially under conditions of inaccurate frame rates and camera rotations.
\section{Discussion}\label{sec4}

This paper proposes a method for 3D trajectory reconstruction of moving points using asynchronous cameras, addressing the challenge of simultaneous trajectory reconstruction and camera synchronization. The experimental results indicate that the proposed method can effectively reconstruct the trajectories of moving points based on asynchronous cameras. Compared with previous iterative algorithms, our algorithm exhibits higher accuracy and efficiency. Especially in cases where the camera frame rate $\alpha_{c}$ is inaccurate, the accuracy of the proposed method is significantly improved. When multiple moving targets are observed, tighter and more continuous constraints on the moving points can be utilized to optimize the cameras' rotation, thereby enhancing the accuracy. The real-world results indicate that the proposed algorithm achieved a localization error of 112.95 m at an observation distance range of 15 $\sim$ 20 km.

This algorithm is applicable to the task of localizing moving targets based on a multi-camera system, which does not require time synchronization among the multiple cameras. When multiple moving targets are observed, tighter and more continuous constraints on the moving points can be utilized to enhance the accuracy. In addition, the proposed algorithm represents the target's motion trajectory using a time polynomial, which holds significant physical meaning. The coefficients of the time polynomial can be used to analyze the kinematic and dynamic characteristics of the target. In summary, the proposed algorithm is applicable to many UAV applications. Future work involves optimizing the UAVs' observation trajectory. By planning the trajectory of the UAV, we can improve the observation conditions, thereby enhancing the measurement accuracy.
\section{Conclusions}\label{sec5}

This paper proposes a novel trajectory reconstruction method of moving points based on asynchronous cameras. Most present methods address only one of the coupled sub-problems: either trajectory reconstruction or camera synchronization. The iterative solution method has deficiencies in both efficiency and accuracy. We propose a bundle adjustment framework that simultaneously addresses these two coupled sub-problems. We also utilize tighter constraints on moving points to optimize camera rotations alongside asynchronous cameras' time information and the target's motion parameters. This method resolves the issue of low measurement accuracy in camera rotations encountered in practical applications, thereby enhancing the accuracy of target trajectory reconstruction. We conduct experiments on both simulated and real-world data. The experimental results demonstrate the feasibility and accuracy of the proposed method. The real-world results indicate that the proposed algorithm achieved a localization error of 112.95 m at an observation distance range of 15 $\sim$ 20 km.

\Acknowledgements{This work was supported by the Hunan Provincial Natural Science Foundation for Excellent Young Scholars (Grants 2023JJ20045) and the National Natural Science Foundation of China (Grant 12372189).}

\end{multicols}

\makeentitle

\end{document}